\documentclass[10pt,twocolumn,letterpaper]{article}

\usepackage[T1]{fontenc}
\usepackage[pagenumbers]{cvpr}
\usepackage{cuted}
\usepackage{multirow}
\usepackage{wrapfig}
\usepackage[table]{xcolor}
\usepackage{tablefootnote}
\usepackage[ruled]{algorithm2e}
\usepackage{listings}
\graphicspath{{images/}}

\definecolor{cvprblue}{rgb}{0.21,0.49,0.74}
\usepackage[pagebackref,breaklinks,colorlinks,allcolors=cvprblue]{hyperref}

\newcommand{\paravspace}{\vspace{-11pt}}

\def\confName{CVPR}
\def\confYear{2026}

\definecolor{G}{rgb}{0.8, 1.0, 0.8}
\definecolor{R}{rgb}{1.0, 0.8, 0.8}

\SetCommentSty{mycommentfont}

\title{Real-Time Generation of Streamable Talking Portrait Video with Reference-Guided Deep Compression VAEs}

\author{Sicheng Xu$^{1}$ \hspace{0.5em} Yu Deng$^{1}$  \hspace{0.5em} Shoukang Hu$^{1}$  \hspace{0.5em} Yichuan Wang$^{2}$  \hspace{0.5em} Yizhong Zhang$^{1}$ \\ Zhan Chen$^{2}$  \hspace{0.5em} Jiaolong Yang$^{1}$ \hspace{0.5em} Baining Guo$^{1}$
\\
$^1${Microsoft Research} \hspace{1em} $^2${Microsoft AI}
}

\begin{document}

\maketitle

\begin{strip}
  \includegraphics[width=\textwidth]{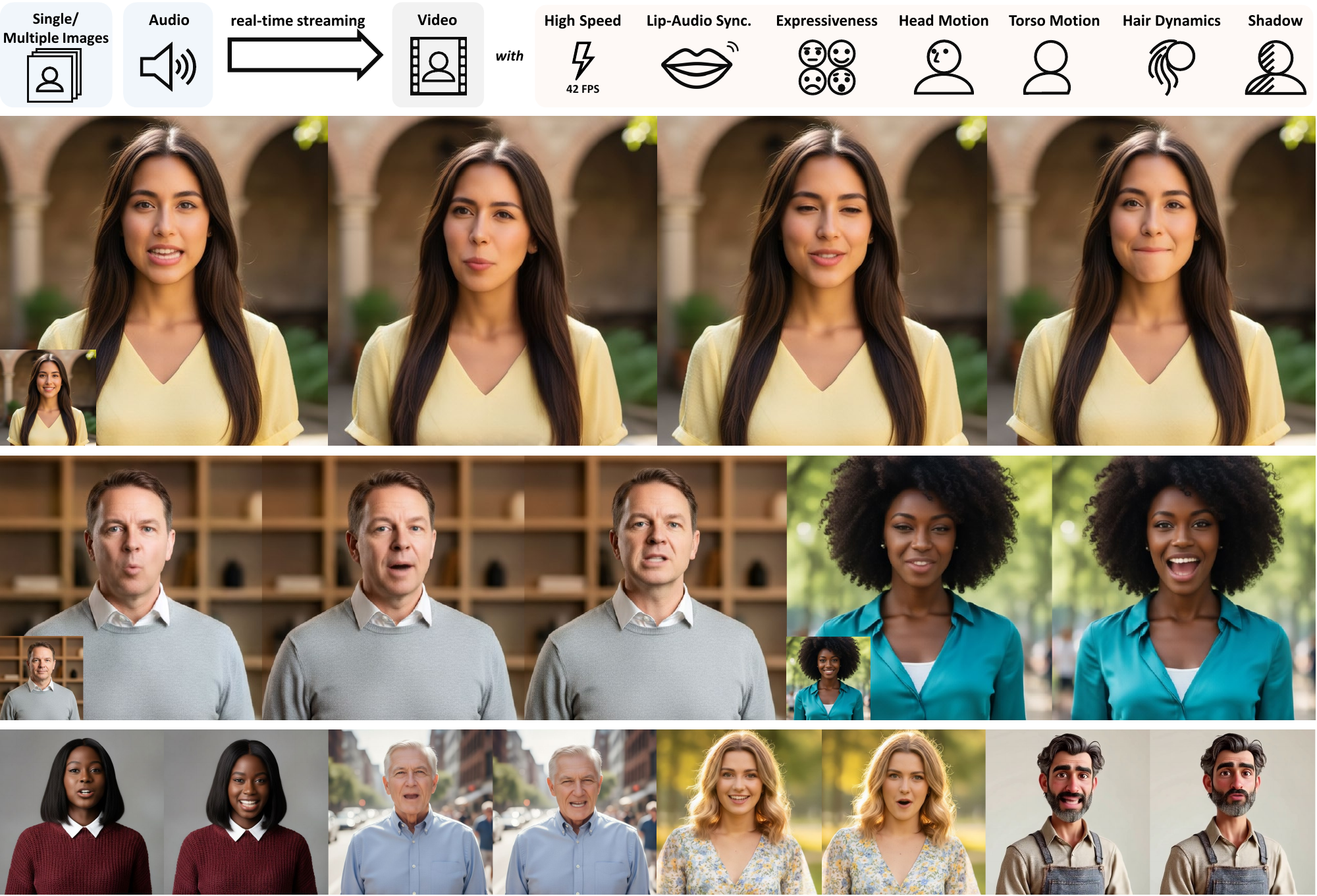}
  \vspace{-10pt}
\captionsetup{type=figure,font=small,position=top}
  \caption{Our method synthesizes streamable  talking portrait videos given speech audio and one or multiple reference images, enabling generation of $\mathbf{512\times512}$ videos at $\mathbf{42}$ FPS on a single GPU. The results exhibit rich visual dynamics, including audio-synchronized lip motion, facial expressions, head and torso movement, hair dynamics, and lighting and shadow effects, advancing the level of realism and liveliness for real-time talking portraits. \textit{Identities presented in the paper are non-existent and created by Gemini 2.5 flash image~\cite{fortin2025nanobanana}.}}
  \label{fig:teaser}
\end{strip}

\begin{abstract}

Video diffusion models have significantly advanced portrait video generation, yet their high computational demands limit their use in interactive applications. This work presents a framework for streamable talking portrait video generation conditioned on speech audio and reference images. 
Designed meticulously for streaming scenarios, it features a causal video VAE for deep latent compression and an autoregressive latent denoising model. 
Our causal VAE integrates a variable number of reference images as guidance, allowing the network to focus on dynamic information rather than static appearance, thereby enhancing compression efficacy and reconstruction quality. Additionally, we extend the residual auto-encoding paradigm to improve spatial-temporal causality handling in our VAE. 
The generator is based on a Rectified Flow Transformer architecture and produces video latents in a blockwise auto-regressive manner. Our method enables the real-time generation of high-quality talking portrait videos, achieving speeds significantly faster than baseline models.  Furthermore, comprehensive experiments demonstrate that it is on par with or even outperforms these large models in realism, vividness, and video quality.

\end{abstract}
\section{Introduction}

With the advent of Large Language Models such as ChatGPT~\cite{2024GPT4o}, there has been a growing trend toward leveraging human–AI interaction tools to enhance work efficiency and interpersonal communication. 
Among these tools, lifelike talking portraits featuring dynamic visual representations play a crucial role in conveying subtle cues that go beyond spoken or written language.
For example, they are essential for applications such as social companionship, interactive education, and mental health support, where dynamic and empathetic communication is paramount. 

Generating talking portrait videos from audio has been actively studied in the past. Most earlier works have focused on the face or head region~\cite{chen2018lip,prajwal2020lip,wang2021audio2head,wang2021one,zhang2023sadtalker,wang2023progressive,xu2024vasa} and highly-realistic talking head results have been achieved~\cite{xu2024vasa,guo2024liveportrait}. However, these methods often rely on specific human face or head priors and extending them to encompass larger portrait regions like the torso is difficult. Additionally, their ability to generate non-rigid dynamics beyond facial areas and complex lighting and shadow effects is limited, preventing them from reaching a higher degree of realism.

Recent advances in large diffusion models for video generation have dramatically pushed the boundaries for talking portrait video~\cite{rombach2022high,blattmann2023stable,stypulkowski2024diffused,tian2024emo,xu2024hallo,wei2024aniportrait,wang2025fantasytalking,lin2025omnihuman}. These large models are able to generate vivid and realistic dynamics across much larger regions beyond head and produce high-quality videos with strong aesthetic appeal.
However, their high computational cost restricts their application to offline content generation, hindering their use in broader real-time and interactive streaming scenarios. For instance, state-of-the-art models require several minutes to generate a 5-second video on modern GPUs~\cite{wang2025fantasytalking,Ji_2025_CVPR,cui2025hallo3}. 
Although various approaches have been proposed to improve the efficiency of video diffusion~\cite{yin2024onestep,yin2024improved}, adapting the large models for real-time streaming scenarios is still a formidable task.

Our goal is to achieve efficient talking portrait video generation from audio. Specifically, we have three objectives: \emph{\textbf{1)}} \emph{real-time, streamable generation} to support interactive applications; \emph{\textbf{2)}} \emph{high visual quality and vividness} to ensure realism, including precise audio-lip synchronization, vivid facial and body motions, realistic lighting and dynamic effects, and high video quality; \emph{\textbf{3)}} \emph{handling a wide torso region beyond the head} to improve immersiveness. Collectively, these requirements present significant challenges, particularly in achieving both efficiency and quality to a high standard, which remains largely unattainable with the existing methods.

Towards these objectives, we present a new framework for streamable talking portrait video generation. Our models include a \emph{causal} video VAE for latent compression and an \emph{autoregressive} latent denoising model conditioned on audio. The causality of the VAE ensures smooth video generation given arbitrary-length video latents, preventing temporal discontinuities. Meanwhile, the autoregressive scheme of the denoising generator which can leverage key–value (KV) caching for efficient inference is naturally suited for streaming.

To obtain a compact latent space for efficient generation, we introduce two key designs for our causal VAE. First, in contrast to ordinary VAEs, we introduce \emph{reference image guidance} to the VAE decoder. 
The intuition is that, unlike generic video generation, talking portrait videos typically feature a fixed subject as the dominant content. The user's reference image shares significant information with the generated videos. By incorporating it into the VAE, compression efficacy is enhanced as the network can focus less on the subject's appearance and background, and more on extracting dynamic information. We train the VAE to accept one or multiple reference images and they significantly improve reconstruction quality.
Second, we extend the residual auto-encoding paradigm of DC-AE~\cite{chen2024deep} to our causal video VAE. Temporal and spatial residual encoding are applied in separate steps when resolution changes, with the first frame handled individually to preserve causality in temporal processing. This design also leads to significantly enhanced reconstruction fidelity.

Given the learned compact latent vectors, we apply a Rectified Flow Transformer~\cite{lipman2022flow,esser2024scaling} as the generator conditioned on the audio input and the reference image(s). 
The generator is trained in a blockwise autoregressive manner with a teacher-forcing strategy~\cite{zhou2025taming}. We train our VAE and generator on a corpus of talking portrait videos.
Comprehensive experiments show that our method is not only significantly faster than the baseline portrait video generation models but also producing comparable or even better results than these models across various metrics of realism and vividness.

Our \textbf{main contributions} are summarized as follows:
\vspace{-1pt}
\begin{itemize}
	\item We present a framework for streamable talking portrait video generation with a block-wise autoregressive denoising model and a causal video decoder, which can continuously generate arbitrary-length video from audio.
	\item We propose a reference-guided video VAE with a causal residual video auto-encoding scheme adapted from DC-AE~\cite{chen2024deep}, which can incorporate one or more reference portrait images into decoder to attain significantly enhanced video reconstruction fidelity.
	\item We implement a video compression rate of 768, approximately \textbf{10-15$\times$ higher} than those of the VAE in popular video diffusion models~\cite{yang2024cogvideox,wan2025}, and we achieve a video generation speed of 42 FPS, over \textbf{25$\times$ faster} than existing diffusion-based talking portrait generation model~\cite{wang2025fantasytalking,Ji_2025_CVPR,cui2025hallo3}, and meanwhile \textbf{comparable or higher quality}  results compared to these models across various metrics. 
\end{itemize}

\section{Related Work}

\paragraph{Audio-driven talking portrait generation.} 
Generating talking head videos from audio input has been a long-standing problem in computer vision and graphics.
Early works primarily focus on synthesizing lip movements synchronized with the input speech~\cite{brand1999voice,bregler2023video,suwajanakorn2017synthesizing,chen2018lip,prajwal2020lip}. Later methods expand the scope to full-head generation, modeling a broader range of motions such as eye gaze and blinks~\cite{zhang2023sadtalker,wang2023progressive}, facial expressions~\cite{yin2022styleheat,liang2022expressive,ji2022eamm,gururani2023space,yu2023talking}, and head pose~\cite{zhou2020makelttalk,wang2021audio2head,zhou2021pose,ye2024real3d}.
These methods typically leverage motion representation learning, where motion is explicitly disentangled from appearance and encoded into compact forms such as sparse keypoints~\cite{qi2025chatanyone, gururani2023space,liang2022expressive}, 3D parameters~\cite{yu2023talking,ye2024real3d}, or learned latent representations~\cite{xu2024vasa,drobyshev2024emoportraits}.
This design simplifies generation complexity and improves controllability, but limits fine-grained realism, making it difficult to capture subtle effects like hair dynamics or dynamics beyond the head region, like torso and chest movement.
Recently, large video generation foundation models~\cite{rombach2022high,yang2024cogvideox,lin2025diffusion,wan2025} have enabled high-fidelity portrait video generation~\cite{tian2024emo,wei2024aniportrait,cui2025hallo3,lin2025omnihuman,chen2025echomimic,li2025tokenmotion,yuan2025identity,wang2025echoshot}.
While these methods excel at capturing rich video dynamics, their high computational cost makes them impractical for real-time applications. 
Our work aims to bridge this gap by achieving fine-grained motion fidelity with real-time generation efficiency.

\paravspace
\paragraph{Efficient image and video generation.}
Score-based generative modeling~\cite{ho2020denoising,song2020score, liu2022flow,lipman2022flow} has emerged as a leading approach for modern image and video generation. A common practice~\cite{rombach2022high, yang2024cogvideox, wan2025} is to employ a VAE to compress visual data into a compact latent space, thus reducing the token context length required for diffusion-based generation. Recent approaches~\cite{chen2024deep,yu2024efficient,hacohen2024ltx,yu2024image,tian2025reducio,chen2025dc,peng2025open} further explore acceleration by increasing the compression ratio, effectively lowering the computational cost during generation. 
However, these methods primarily focus on static image synthesis or offline video generation. Their applications to real-time, streamable talking portrait generation, which demands both low-latency and temporally coherent synthesis, still underexplored.

Instead of increasing the VAE compression ratio to reduce latent context, another common approach uses post-training distillation~\cite{sauer2024adversarial,sauer2024fast,yin2024one,yin2024improved} to accelerate generation by decreasing the number of denoising steps. Some recent methods further combine autoregressive modeling~\cite{lin2025autoregressive, yin2025slow, low2025talkingmachines} with distillation to enable real-time video generation. 
In contrast, our method achieves real-time $512{\times}512$ talking video generation without distillation, thanks to reference-guided deep-compression VAE and a native auto-regressive design that reduces latent token length while supporting sequential generation.

\paravspace
\paragraph{Auto-regressive video generation.} 
Auto-regressive (AR) modeling has gained significant attention in video generation for its ability to capture temporal dependencies while supporting efficient, low-latency, long-horizon synthesis via KV caching. Early works such as PixelRNN and PixelCNN~\cite{van2016conditional, van2016pixel} demonstrated the effectiveness of autoregressive density estimation, inspiring extensions to spatio-temporal generation. Transformer-based AR models on discrete VQ-VAE tokens, such as VideoGPT~\cite{yan2021videogpt}, enabled video synthesis in a compact discrete latent space. Recently, autoregressive mechanisms have also been integrated into diffusion frameworks~\cite{chen2024diffusion, huang2025self,yin2025slow}. MAGI-1~\cite{teng2025magi} further scales AR generation to 24 billion parameters, supporting long-context video synthesis with streaming decoding. In this paper, we also employ auto-regressive modeling with block-wise causal attention, trained using teacher forcing. This design provides an essential and efficient mechanism for streamable real-time talking video generation.

\section{Method}

\begin{figure*}[t]
	\small
	\centering
	\includegraphics[width=\textwidth]{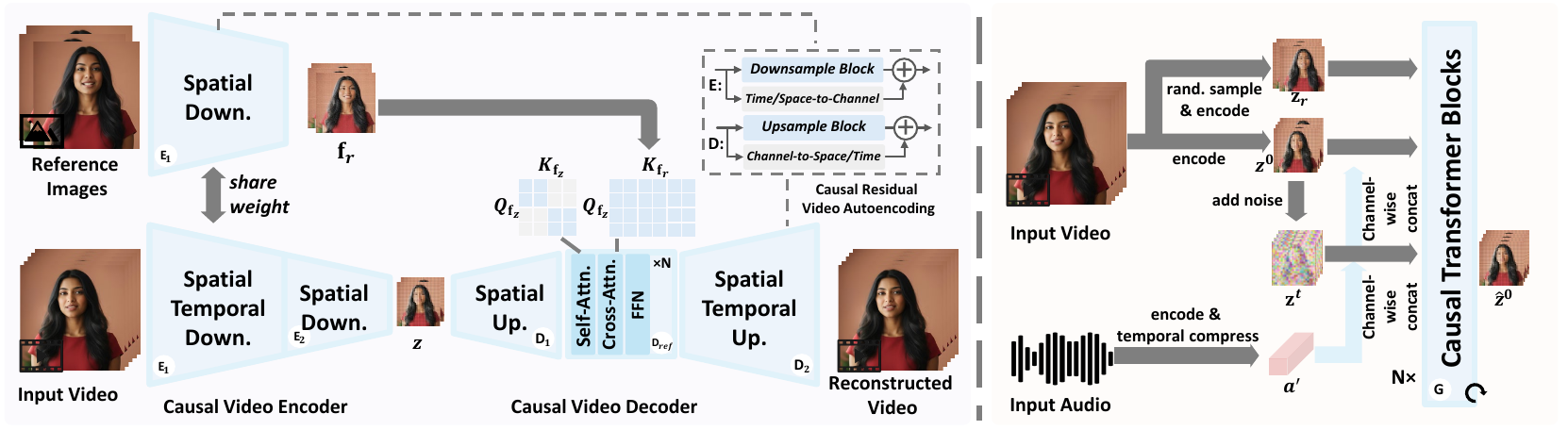}
    \vspace{-18pt}
	\caption{Overview of our framework. \textbf{Left}: The proposed reference-guided causal video VAE. $Q_{\mathbf{f}_z}$ and $K_{\mathbf{f}_z}$ are query–key pairs from $\mathbf{f}_z$, upsampled from video latent $\mathbf{z}$ via $D_1$. \textbf{Right}: Rectified flow transformer with block-wise causal attention for modeling the probabilistic distribution of the compact video latents.}
	\label{fig:framework}
\end{figure*}

Our goal is to generate a realistic talking portrait video from one or several reference images and arbitrary speech audio in real time.
This task can be formulated as a conditional probabilistic modeling problem of the portrait video $\mathbf{y} = [\mathbf{x}_1, \mathbf{x}_2, ..., \mathbf{x}_n ]$, given the reference image set $\mathbf{r}=\{\mathbf{x}_1, \mathbf{x}_2, ..., \mathbf{x}_m\}$ and the audio signal $\mathbf{a}$:
\begin{equation}
\mathbf{y} \sim p(\mathbf{y} \mid \mathbf{r}, \mathbf{a}).
\end{equation}

Following the modern video generation paradigm, we decompose the generation process into two sub-tasks: (1) generating a compact latent representation $\mathbf{z}$ conditioned on the audio, and (2) decoding $\mathbf{z}$ into the final video $\mathbf{y}$.

\subsection{Reference-Guided Deep Video Compression}\label{sec:latent}
For efficient generation, we require a highly compact latent space to reduce the latent token length. We also need to enable streaming video decoding to prevent temporal discontinuity and minimize latency during real-time synthesis.

To this end, we employ a \emph{causal} VAE for latent space learning.
As shown in Fig.~\ref{fig:framework}, our causal auto-encoder adopts a two-stage symmetric architecture. The encoder comprises two cascaded modules: $E_1$ performs both spatial and temporal downsampling, while $E_2$ further compresses the spatial resolution. They compress the input video $\mathbf{y}\in \mathbb{R}^{3 \times(T{+}1)\times H\times W}$ \footnote{The temporal dimension $T+1$ arises from the common design that makes causal video VAEs applicable to a single image (\ie, the first frame).} into a latent tensor $\mathbf{z}\in \mathbb{R}^{C_z\times T_z\times H_z\times W_z}$. The decoder mirrors this hierarchy with $D_1$ handling spatial upsampling and $D_2$ jointly performing spatial–temporal upsampling for reconstruction. $E_1$, $E_2$, $D_1$, $D_2$ are \emph{convolutional neural networks} equipped with \emph{causal convolutions}. We also adopt RMSNorm~\cite{zhang2019} as normalization layer to preserve temporal causality~\cite{wan2025}.

\paravspace
\paragraph{Reference guidance injection with Transformer.}
Unlike generic video generation, talking portrait videos typically feature a fixed subject as the dominant content. The reference images share significant information with the videos. In light of this, we inject reference images into our VAE to improve compression efficacy and enhance reconstruction fidelity. With reference guidance, the network can focus less on the subject's appearance and background content, and more on extracting human motion information. 

Specifically, between these two decoding stages of $D_1$ and $D_2$, we incorporate a transformer-based fusion network $D_{\text{ref}}$ to integrate reference image features and refine the decoded representations.
For reference images $\mathbf{r}\in \mathbb{R}^{C\times M\times  H\times W }$ \footnote{For simplicity, we slightly abuse dimension notation of tensors here: $M$ is used as the batch dimension processed by convolution operations instead of the temporal dimension.}, which are randomly sampled from the input video $\mathbf{y}$ during training or provided by user during inference, we reuse $E_1$ to process $\mathbf{r}$ frame by frame, producing mid-level feature maps $\mathbf{f}_c\in \mathbb{R}^{C_f\times M\times H_f\times W_f}$ which retain appearance cues of the subject and background. The decoder $D_1$ upsamples the latent $\mathbf{z}$ to $\mathbf{f}_z\in \mathbb{R}^{C_f\times T_z \times H_f\times W_f}$ which has the same spatial dimensions as $\mathbf{f}_c$.
Within the transformer fusion module $D_{ref}$, frame-wise self-attention is applied to $\mathbf{f}_z$ to maintain temporal causality, followed by cross-attention layers that inject fine-grained visual information from $\mathbf{f}_c$. The fused features are then passed to $D_2$.

During training, we randomly sample different numbers of reference images to allow the model to handle a variable number of reference inputs provided by the user at test time.

\begin{figure}[t]
	\small
	\centering
	\includegraphics[width=\linewidth]{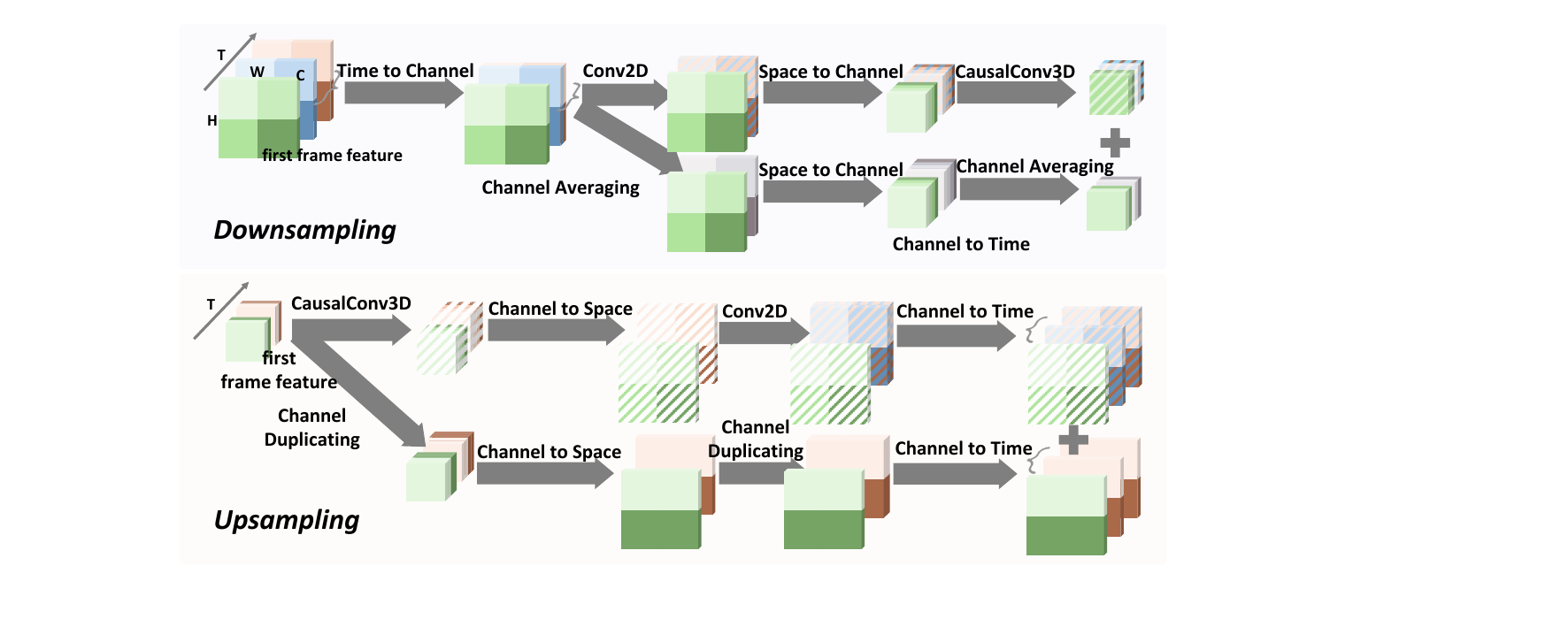}
    \vspace{-17pt}
	\caption{Causal residual video auto-encoding. We apply separate temporal and spatial down/up-sampling with residual encoding. The first frame is handled independently to preserve causality. }
	\label{fig:crva}
\end{figure}

\paravspace
\paragraph{Causal residual video auto-encoding (CR-VA).} To further enhance reconstruction quality under high compression rates, we extend the residual auto-encoding paradigm of DC-AE~\cite{chen2024deep} to our causal video VAE. 
Specifically, for resolution changes in the VAE, we apply temporal and spatial residual encoding in a two-step process:
\begin{enumerate}[label=\arabic*)]
	\item \emph{Temporal residual} (if applicable): apply channel-to-time/time-to-channel to change temporal resolution, with channel averaging/duplication for dimension matching. The first frame is excluded.
	
	\item \emph{Spatial residual}: 
	apply channel-to-space/space-to-channel to all frames, also using channel averaging/duplication for dimension matching. This step is applied after temporal processing when enabled.
\end{enumerate}
The main branch applies the same temporal and spatial operations as the residual encoding, but uses convolution layers for feature dimension matching. Figure~\ref{fig:crva} illustrates the downsampling and upsampling process. More detailed explanations can be found in the supplementary material.

\paravspace
\paragraph{VAE Training and compression ratio.}
The training process uses the objective to minimize the discrepancy between the reconstructed frames $\hat{\mathbf{y}}$ and the original input $\mathbf{y}$:

\begin{equation}
	\mathbb{E}_{\hat{\mathbf{y}}} \left[ \lambda_1 \| \hat{\mathbf{y}} - \mathbf{y} \|_1 + \lambda_2 \text{LPIPS}(\hat{\mathbf{y}}, \mathbf{y}) \right], \label{eq:loss}
\end{equation}
where $\text{LPIPS}$ measures perceptual difference~\cite{zhang2018unreasonable}, and $\lambda_1$, $\lambda_2$ are scalar weights. We also incorporate a KL-regularization on $\mathbf{z}$ to encourage a well-structured latent space following~\cite{rombach2022high,kingma2013auto}.

Our VAE encodes $\mathbf{y} \in \mathbb{R}^{3 \times(T{+}1)\times H\times W}$ with spatial and temporal downsampling factors of $64$ and $4$, respectively, and a latent channel dimension of $64$. 
The resultant latent tensor $\mathbf{z} \in \mathbb{R}^{C_z\times T_z\times H_z\times W_z}$ has dimensions $C_z$, $T_z$, $H_z$, and $W_z$ equal to $64$, $(T/4{+}1)$, $H/64$, and $W/64$, respectively. This amounts to an \emph{overall compression ratio of 768}, which is significantly higher than those of popular generic video generation models (\eg, $48$ in \cite{yang2024cogvideox,wan2025}).

\subsection{
Blockwise AutoRegressive Latent Generation
}\label{sec:generation}
For latent generation, we model the distribution of $\mathbf{z}$ conditioned on the reference image set $\mathbf{r}$ and audio $\mathbf{a}$.
We adopt a Rectified Flow framework~\cite{liu2022flow,lipman2022flow}, which formulates the generation process as solving an ODE that models the transport vector field from a noise distribution to the target data distribution. We apply a Transformer network $G$ to approximate the conditional velocity field with audio–visual inputs in an autoregressive manner, as described below.

\paravspace
\paragraph{Input encoding.}
We encode $\mathbf{r}$ in a frame-wise manner using the encoder to obtain the reference latents $\mathbf{z}_r$ as visual conditions.
For the audio input, we apply a pretrained audio encoder~\cite{prajwal2020lip} to extract audio features, which are further compressed by a trainable temporal embedder (by a factor of $4$) to temporally align with the video latents.
The compressed audio features are then broadcast along the spatial axes, matching the spatiotemporal dimensions of $\mathbf{z}$ for element-wise fusion, yielding $\mathbf{a}'$. To construct the input to $G$ for training, we encode video clips into latent representations $\mathbf{z}$, which serve as the generation targets. Then we add noise $\boldsymbol{\epsilon}^t$ to $\mathbf{z}$ to construct noised latent $\mathbf{z}_t$ for each time step $t$. 
The noisy latent $\mathbf{z}^t$ is concatenated with $\mathbf{a}'$ along the channel dimension. The resultant tensor is then flattened and concatenated with the flattened reference latents $\mathbf{z}_r$, forming the input to the network $G$. The denoising timestep $t$ is injected via AdaLN~\cite{peebles2023scalable}.

\paravspace
\paragraph{Blockwise attention and autoregressive generation.}

\begin{wrapfigure}{r}{0.2\textwidth}
  \centering
  \vspace{0pt}
  \includegraphics[width=0.19\textwidth]{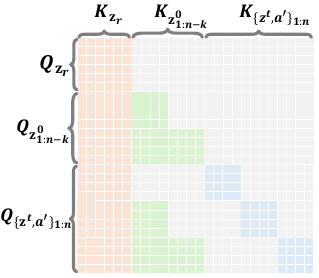}
  \vspace{-8pt}
    \caption{The causal blockwise attention mask.\label{fig:attention}}
  \vspace{-8pt}
\end{wrapfigure} 
To facilitate streamable generation and reduce latency, we configure $G$ to generate latents in an autoregressive manner. In this setup, the content to be generated at each time only attends to the generated context in the past, thus preserving temporal causality. 
To improve computational efficiency, we partition the latent sequence into non-overlapping blocks of size $k$ for blockwise generation. As illustrated in Fig.~\ref{fig:attention}, full self-attention is applied for latents within each block, while inter-block attention is restricted to preceding blocks only. This blockwise causal attention maintains the temporal causality enforced by the autoregressive design. It reduces memory and computation costs significantly and facilitates streamable generation.

\paravspace
\paragraph{Generator Training.}
We apply the \emph{teacher-forcing} strategy~\cite{williams1989learning} for the autoregressive generation training. To provide clean temporal context, the model is conditioned on the ground-truth of the previous latents during training, augmented with Gaussian noise to prevent drift and reduce train–inference mismatch.

Given the constructed audio–visual inputs, the model is trained via the conditional flow-matching loss:
\begin{equation}
\mathbb{E}_{t,\mathbf{z}^0,\mathbf{\epsilon}, \mathbf{z}_r,\mathbf{a'}}\|G(\mathbf{z}^t, \mathbf{z}_r, \mathbf{a'}, t)-(\boldsymbol{\epsilon}^t-\mathbf{z}^0)
\|^2_2, \label{eq:flow_loss}
\end{equation}
where $\mathbf{z}^0$ is the clean video latent. During training, we randomly replace $\mathbf{z}_r$ and $\mathbf{a'}$ with learnable null embeddings to enable classifier-free guidance (CFG) at inference time. Additionally, $\mathbf{z}_r$ is randomly masked to enable inference with a flexible number of reference images. 
In practice, we randomly sample latent windows as training targets.
Since the first frame is temporally uncompressed and differs from the rest, a learnable mask is added to the first frame if the sampled segment includes it.

\paravspace
\paragraph{Streaming generation and decoding.} At inference time, the model autoregressively predicts video latents block by block, which are streamed to the decoder for video frame generation. Within each window, KV caching is applied to reuse past context and reduce inference costs. For videos exceeding one window, the last block from the previous window is reused as the context for the next, ensuring seamless window-by-window generation.

\section{Experiments}

\begin{figure*}[t]
   \begin{center}
      \includegraphics[width=0.85\linewidth]{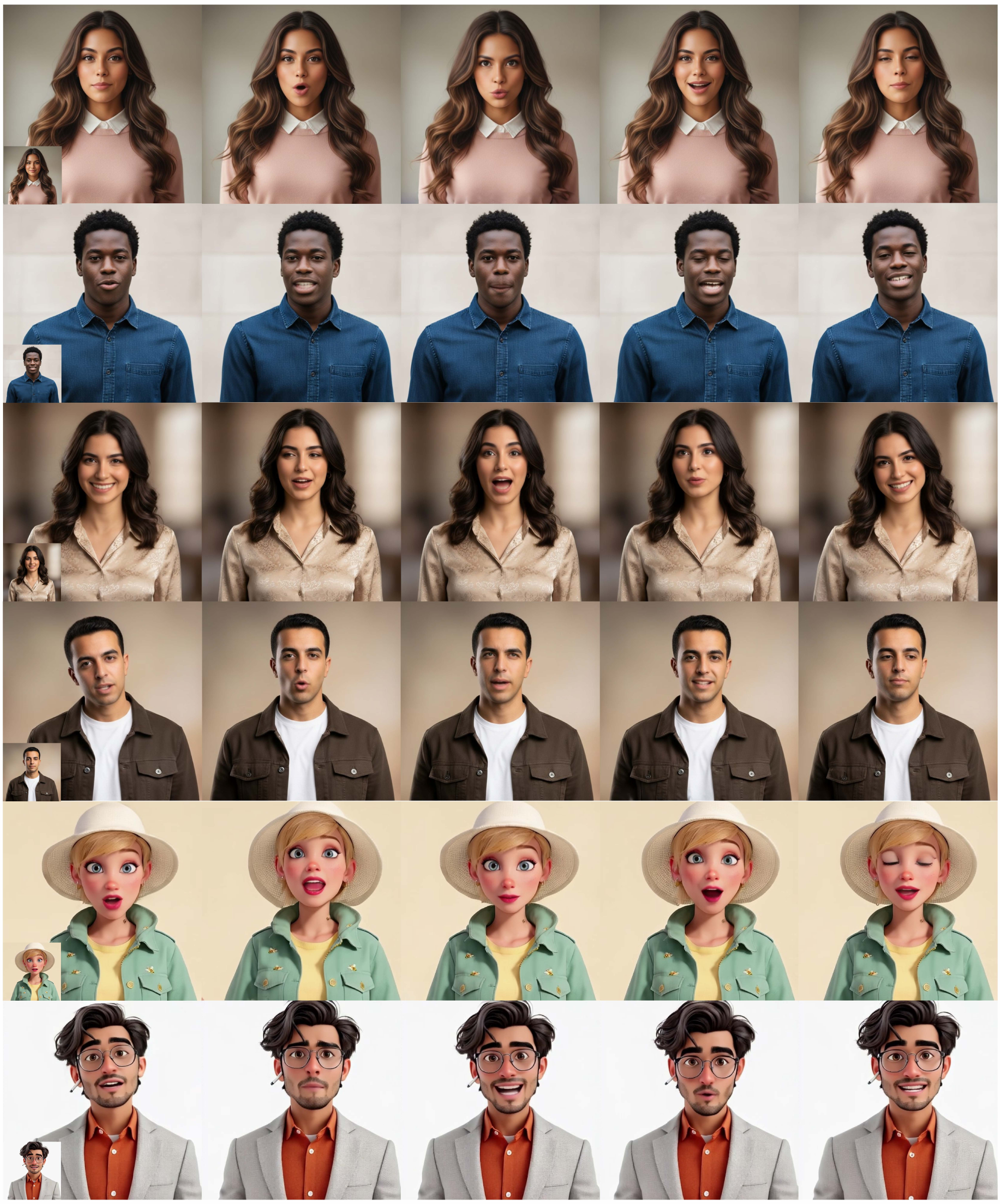}
   \end{center}
   \vspace{-20pt}
      \caption{Talking portrait videos generated by our method, which naturally and comprehensively capture portrait video dynamics. \textit{(Best viewed with zooming in; see the supplementary videos for a more comprehensive evaluation.)}}
   \label{fig:results}
   \vspace{-10pt}
\end{figure*}

\begin{table*}[t!]
	\centering
	\caption{Quantitative results of audio-driven portrait video generation on two benchmarks at ${512\times512}$ resolution. Inference speed is measured on a single H100 GPU. $M$ denotes the number of reference images used; all other methods and ours with $M=1$ use the first frame as reference for generation. FPS is calculated as the number of frames per generation window of  each method, divided by its generation time (we use a window with the longest KV cache to report the speed lower bound of our method). 
    FantasyTalking's results on PortraitOneMin are not reported as its implementation failed to handle long video generation.}
    \vspace{-6pt}
	{\footnotesize
		\begin{tabular}{c@{}c|cccc|cccc|c} 
			\toprule
           \multicolumn{2}{c|}{} &  \multicolumn{4}{c|}{HDTF} & \multicolumn{4}{c|}{PortraitOneMin} & Speed \\
            \midrule
			 \multicolumn{2}{c|}{Method} &$S_C\uparrow$ & $S_D\downarrow$& \!CAPP$\uparrow$ &$\!\!\text{FVD}_{25}\downarrow$\!\! &$S_C\uparrow$ & $S_D\downarrow$& \!CAPP$\uparrow$\! & $\text{FVD}_{25}\downarrow$ & FPS $\uparrow$\\
			\midrule
               \multicolumn{2}{c|}{EchoMIMIC} & \cellcolor[rgb]{1.00,0.84,0.70}5.291 & \cellcolor[rgb]{1.00,0.89,0.70}9.557 & \cellcolor[rgb]{1.00,0.82,0.70}0.341 & \cellcolor[rgb]{1.00,0.84,0.70}143.620 & \cellcolor[rgb]{1.00,0.70,0.70}4.863 & \cellcolor[rgb]{1.00,0.70,0.70}9.594 & \cellcolor[rgb]{1.00,0.71,0.70}0.208 & \cellcolor[rgb]{1.00,0.82,0.70}177.141 &1.4 \\

               \multicolumn{2}{c|}{EchoMIMIC-Distilled} &\cellcolor[rgb]{1.00,0.87,0.70}5.513 &\cellcolor[rgb]{1.00,0.92,0.70}9.350 &\cellcolor[rgb]{1.00,0.83,0.70}0.348 &\cellcolor[rgb]{1.00,0.70,0.70}174.061 &\cellcolor[rgb]{1.00,0.81,0.70}5.531 &\cellcolor[rgb]{1.00,0.78,0.70}9.187 &\cellcolor[rgb]{1.00,0.71,0.70}0.201 &\cellcolor[rgb]{1.00,0.70,0.70}201.899 & 13.3\\
               \multicolumn{2}{c|}{Hallo}& \cellcolor[rgb]{0.90,1.00,0.70}7.457 & \cellcolor[rgb]{0.90,1.00,0.70}7.841 & \cellcolor[rgb]{1.00,0.70,0.70}0.242 & \cellcolor[rgb]{0.92,1.00,0.70}90.880 & \cellcolor[rgb]{1.00,1.00,0.70}6.721 & \cellcolor[rgb]{1.00,0.97,0.70}8.275 & \cellcolor[rgb]{1.00,0.72,0.70}0.210 & \cellcolor[rgb]{1.00,0.81,0.70}179.072 &1.2 \\

               \multicolumn{2}{c|}{Hallo2} & \cellcolor[rgb]{0.89,1.00,0.70}7.547 & \cellcolor[rgb]{0.90,1.00,0.70}7.819 & \cellcolor[rgb]{1.00,0.71,0.70}0.247 & \cellcolor[rgb]{0.91,1.00,0.70}88.170 & \cellcolor[rgb]{0.98,1.00,0.70}6.829 & \cellcolor[rgb]{1.00,0.97,0.70}8.244 & \cellcolor[rgb]{1.00,0.70,0.70}0.196 & \cellcolor[rgb]{1.00,0.88,0.70}162.672 & 1.2 \\

               \multicolumn{2}{c|}{Hallo3}& \cellcolor[rgb]{0.92,1.00,0.70}7.256 & \cellcolor[rgb]{0.99,1.00,0.70}8.596 & \cellcolor[rgb]{1.00,0.81,0.70}0.337 & \cellcolor[rgb]{0.85,1.00,0.70}76.430 & \cellcolor[rgb]{0.98,1.00,0.70}6.836 & \cellcolor[rgb]{1.00,0.88,0.70}8.724 & \cellcolor[rgb]{1.00,0.83,0.70}0.301 & \cellcolor[rgb]{1.00,0.82,0.70}175.340 & 0.27 \\
   
               \multicolumn{2}{c|}{Sonic}& \cellcolor[rgb]{0.73,1.00,0.70}8.799 & \cellcolor[rgb]{0.75,1.00,0.70}6.602 & \cellcolor[rgb]{0.76,1.00,0.70}0.689 & \cellcolor[rgb]{0.70,1.00,0.70}43.920 &\cellcolor[rgb]{0.76,1.00,0.70}8.185  &\cellcolor[rgb]{0.78,1.00,0.70}7.031  &\cellcolor[rgb]{0.80,1.00,0.70}0.598  & \cellcolor[rgb]{0.80,1.00,0.70}95.047 & 1.7 \\
              \multicolumn{2}{c|}{FantasyTalking}   & \cellcolor[rgb]{1.00,0.70,0.70}4.167 & \cellcolor[rgb]{1.00,0.70,0.70}11.144 & \cellcolor[rgb]{1.00,0.90,0.70}0.407 & \cellcolor[rgb]{0.91,1.00,0.70}89.726 &  \multicolumn{4}{c|}{---} &0.36 \\
               \midrule
               \multirow{3}{*}{Ours} 
               & $M=1$  & \cellcolor[rgb]{0.71,1.00,0.70}8.943 & \cellcolor[rgb]{0.71,1.00,0.70}6.286 & \cellcolor[rgb]{0.75,1.00,0.70}0.699 & \cellcolor[rgb]{0.79,1.00,0.70}62.300 &\cellcolor[rgb]{0.70,1.00,0.70}8.537  &\cellcolor[rgb]{0.70,1.00,0.70}6.619  &\cellcolor[rgb]{0.74,1.00,0.70}0.648  & \cellcolor[rgb]{0.79,1.00,0.70}91.964 &\multirow{3}*{42.3} \\

               & $M=2$ & \cellcolor[rgb]{0.70,1.00,0.70}9.056 & \cellcolor[rgb]{0.70,1.00,0.70}6.175 & \cellcolor[rgb]{0.70,1.00,0.70}0.739 & \cellcolor[rgb]{0.73,1.00,0.70}49.400 & \cellcolor[rgb]{0.72,1.00,0.70}8.438 & \cellcolor[rgb]{0.71,1.00,0.70}6.688 & \cellcolor[rgb]{0.70,1.00,0.70}0.677 & \cellcolor[rgb]{0.74,1.00,0.70}81.517 &  \\

               & $M=3$ & \cellcolor[rgb]{0.71,1.00,0.70}8.998 & \cellcolor[rgb]{0.71,1.00,0.70}6.226 & \cellcolor[rgb]{0.70,1.00,0.70}0.739 & \cellcolor[rgb]{0.70,1.00,0.70}43.270 &\cellcolor[rgb]{0.70,1.00,0.70}8.546  &\cellcolor[rgb]{0.71,1.00,0.70}6.648  &\cellcolor[rgb]{0.72,1.00,0.70}0.659  & \cellcolor[rgb]{0.70,1.00,0.70}73.693 \\
			\bottomrule
		\end{tabular}
	}
	\label{table:portrait_gen}
\end{table*}

\begin{table*}[t!]
\centering
\caption{
Ablation study on the effects of reference guidance and video residual auto-encoding (VRA) for our VAE. $M$ denotes the number of reference images, and $\Delta\text{PSNR}$ indicates the PSNR improvement obtained by using reference guidance. 
}
\vspace{-6pt}
{\footnotesize

\begin{tabular}{cl|cccc|cccc}
\toprule
\multicolumn{2}{c|}{} & \multicolumn{4}{c|}{VoxCeleb2} & \multicolumn{4}{c}{HDTF} \\
\midrule
\multicolumn{2}{c|}{Configuration} & $L_1 \downarrow$ & PSNR $\uparrow$ & $\Delta$PSNR & LPIPS $\downarrow$ & $L_1 \downarrow$ & PSNR $\uparrow$ & $\Delta$PSNR & LPIPS $\downarrow$ \\
\midrule
\multirow{4}{*}{w/o CR-VA} & $M=0$ {\scriptsize{(No ref.)}} & \cellcolor[rgb]{1.00,0.70,0.70}0.020 & \cellcolor[rgb]{1.00,0.70,0.70}29.071 & -- & \cellcolor[rgb]{1.00,0.70,0.70}0.088 & \cellcolor[rgb]{1.00,0.70,0.70}0.021 & \cellcolor[rgb]{1.00,0.70,0.70}28.306 & -- & \cellcolor[rgb]{1.00,0.70,0.70}0.087 \\
& $M=1$ & \cellcolor[rgb]{0.90,1.00,0.70}0.014 & \cellcolor[rgb]{0.98,1.00,0.70}31.676 & \cellcolor[rgb]{1.00,0.70,0.70}2.605 & \cellcolor[rgb]{0.91,1.00,0.70}0.051 & \cellcolor[rgb]{0.87,1.00,0.70}0.013 & \cellcolor[rgb]{0.98,1.00,0.70}32.068 & \cellcolor[rgb]{1.00,0.70,0.70}3.762 & \cellcolor[rgb]{0.86,1.00,0.70}0.040 \\
& $M=2$ & \cellcolor[rgb]{0.84,1.00,0.70}0.013 & \cellcolor[rgb]{0.91,1.00,0.70}32.305 & \cellcolor[rgb]{1.00,0.91,0.70}3.234 & \cellcolor[rgb]{0.83,1.00,0.70}0.043 & \cellcolor[rgb]{0.81,1.00,0.70}0.012 & \cellcolor[rgb]{0.93,1.00,0.70}32.663 & \cellcolor[rgb]{1.00,0.82,0.70}4.357 & \cellcolor[rgb]{0.82,1.00,0.70}0.035 \\
& $M=3$ & \cellcolor[rgb]{0.77,1.00,0.70}0.012 & \cellcolor[rgb]{0.85,1.00,0.70}32.766 & \cellcolor[rgb]{0.93,1.00,0.70}3.695 & \cellcolor[rgb]{0.79,1.00,0.70}0.039 & \cellcolor[rgb]{0.81,1.00,0.70}0.012 & \cellcolor[rgb]{0.89,1.00,0.70}33.149 & \cellcolor[rgb]{1.00,0.92,0.70}4.843 & \cellcolor[rgb]{0.79,1.00,0.70}0.032 \\
\midrule
 & $M=0$ {\scriptsize{(No ref.)}} & \cellcolor[rgb]{1.00,0.83,0.70}0.018 & \cellcolor[rgb]{1.00,0.76,0.70}29.604 & -- & \cellcolor[rgb]{1.00,0.71,0.70}0.087 & \cellcolor[rgb]{1.00,0.75,0.70}0.020 & \cellcolor[rgb]{1.00,0.73,0.70}28.678 & -- & \cellcolor[rgb]{1.00,0.71,0.70}0.086 \\
w. CR-VA & $M=1$ & \cellcolor[rgb]{0.84,1.00,0.70}0.013 & \cellcolor[rgb]{0.90,1.00,0.70}32.354 & \cellcolor[rgb]{1.00,0.75,0.70}2.750 & \cellcolor[rgb]{0.85,1.00,0.70}0.045 & \cellcolor[rgb]{0.81,1.00,0.70}0.012 & \cellcolor[rgb]{0.87,1.00,0.70}33.469 & \cellcolor[rgb]{1.00,0.91,0.70}4.791 & \cellcolor[rgb]{0.79,1.00,0.70}0.032 \\
(Ours) & $M=2$ & \cellcolor[rgb]{0.77,1.00,0.70}0.012 & \cellcolor[rgb]{0.79,1.00,0.70}33.281 & \cellcolor[rgb]{0.94,1.00,0.70}3.677 & \cellcolor[rgb]{0.76,1.00,0.70}0.036 & \cellcolor[rgb]{0.71,1.00,0.70}0.010 & \cellcolor[rgb]{0.78,1.00,0.70}34.510 & \cellcolor[rgb]{0.88,1.00,0.70}5.832 & \cellcolor[rgb]{0.74,1.00,0.70}0.027 \\
& $M=3$ & \cellcolor[rgb]{0.71,1.00,0.70}0.011 & \cellcolor[rgb]{0.71,1.00,0.70}33.979 & \cellcolor[rgb]{0.71,1.00,0.70}4.375 & \cellcolor[rgb]{0.71,1.00,0.70}0.031 & \cellcolor[rgb]{0.71,1.00,0.70}0.010 & \cellcolor[rgb]{0.71,1.00,0.70}35.374 & \cellcolor[rgb]{0.71,1.00,0.70}6.696 & \cellcolor[rgb]{0.71,1.00,0.70}0.023 \\

\bottomrule
\end{tabular}
}
\label{table:ablation_residual_ref}
\end{table*}

\subsection{Implementation Details}

\paragraph{Network architectures.}  
In our causal video VAE, the four modules $E_1$, $E_2$, $D_1$, $D_2$ are primarily built on causal convolutional residual blocks. $E_1$ contains eight residual blocks, where every two blocks perform spatial downsampling, resulting in a total spatial compression factor of 16. 
Temporal downsampling is applied at the 4th and 6th blocks, leading to a temporal compression factor of 4. 
The output feature dimension of $E_1$ is 1024. 
$E_2$ further compresses the latent with two residual blocks, each performing one additional downsampling. 
The decoders $D_1$ and $D_2$ are symmetric to $E_2$ and $E_1$, respectively. 
The reference-guidance decoder $D_{\text{ref}}$ consists of six transformer blocks with a head dimension of 64 and 16 attention heads.

For the video generator $G$, two MLP-based audio embedders are used to temporally align audio features with video latents, one dedicated to the first audio frame and the other shared across the remaining frames. 
$G$ comprises 24 transformer blocks with 12 attention heads and a head dimension of 128. 
We employ 3D Rotary Positional Embedding (3D RoPE)~\cite{wan2025} to encode the spatiotemporal positions of video latents, ensuring consistent temporal modeling. Ground-truth latents serving as context are augmented with noise levels from $t=0$ to $t=0.7$.

\paravspace
\paragraph{Training and inference details.}
We train our model on filtered VoxCeleb2~\cite{chung2018voxceleb2} data that contain 50 hours of talking portrait videos and an in-house talking portrait dataset with 280 hours of videos. The training set contains about 10K unique identities. The video frames are cropped to portrait regions. We segment the videos into clips with lengths of up to 10 seconds for training.

For efficient training, the VAE is initially trained on 5-frame clips at $256^2$ resolution. The clip length is progressively increased during training to reduce temporal drift and enhance temporal consistency. The model is then fine-tuned at a higher resolution of $512^2$.
For the generator $G$, one generation window comprises 32 generated latent frames, corresponding to 128 video frames (or 125 for the first window). The block size $k$ is set to 4 latent frames.
During training, the number of reference conditions is randomly sampled between 1 and 3.
At inference, the same reference images are used for both generation and VAE decoding. Classifier-free guidance is applied to audio features and reference latents (scale = 2). The denoising process uses 12 steps with a timestep shift~\cite{esser2024scaling} of 5, set empirically.
More details can be found in the supplementary material.

\paravspace

\paragraph{Evaluation benchmarks.} 
We evaluate our method on two benchmarks. 
The first is HDTF~\cite{zhang2021flow}, processed in the same manner as the training data. 
To ensure a balanced representation across identities, we randomly sample up to two segments per identity, resulting in a total of 123 segments from 66 unseen identities. The second benchmark is a self-collected dataset called PortraitOneMin, which consists of 32 one-minute clips from 16 unseen identities, sourced from online coaching and educational lectures, featuring diverse speaking styles.
For the video VAE reconstruction ablation, we include a test split from VoxCeleb2 containing approximately 1K clips. All evaluations are conducted at 512$\times$512 resolution.

\subsection{Talking Portrait Generation Results}

Figure~\ref{fig:teaser} and~\ref{fig:results} showcase talking portrait videos generated by our method. The identities are synthetic and non-existent, created by Gemini 2.5 Flash image~\cite{fortin2025nanobanana} based on text prompts provided by GPT-4o~\cite{2024GPT4o}. Our method generates high-quality portrait videos that are both realistic and lifelike, featuring vivid dynamics such as accurate lip-audio synchronization, expressive facial expressions, natural head and torso movements, subtle hair dynamics, shadow effects, etc. \textit{We highly recommend readers watch the videos in the supplementary material to better appreciate the quality, naturalness, and vividness of our results.}

\subsection{Comparison with Prior Methods}

\subsubsection{Audio-Driven Talking Portrait Generation} Recent audio-driven talking portrait approaches are predominantly offline methods leveraging large image or video diffusion models. We compare our method with several representative approaches, including Echomimic and its distilled version~\cite{chen2025echomimic}, the Hallo series~\cite{xu2024hallo,cuihallo2,cui2025hallo3}, Sonic~\cite{Ji_2025_CVPR}, and FantasyTalking~\cite{wang2025fantasytalking}. Echomimic and Hallo are built on Stable Diffusion~\cite{rombach2022high}; Hallo2 and Sonic are based on Stable Video Diffusion~\cite{blattmann2023stable}; Hallo3 leverages CogVideoX~\cite{yang2024cogvideox}; and FantasyTalking builds upon Wan2.1~\cite{wan2025}. \textit{Note that these methods require significantly more computational resources and are unable to perform real-time generation}. The comparison with these methods primarily provides a reference for our generation quality. 

\paravspace
\paragraph{Quantitative evaluations.}
Table~\ref{table:portrait_gen} compares our method with previous approaches on HDTF and PortraitOneMin.
For audio-lip synchronization, we use the confidence score $S_C$ and the feature distance $S_D$ from SyncNet~\cite{chung2017out} as metrics. For head pose, we use the CAPP metric proposed by~\cite{xu2024vasa} to measure its alignment with the input audio. To assess overall video quality, we compute the Fréchet Video Distance (FVD)~\cite{unterthiner2019fvd} using sequences of 25 consecutive frames.  
All compared methods use the first frame as condition.
Since our method supports multi-reference  inference, we include two additional variants: for $M=2$, the last frame of the original video is added as a reference; for $M=3$, the middle frame is also included.

For the single-reference setting, our method achieves the best lip-sync and head-audio alignment. It provides comparable or better overall video quality on both benchmarks, even against large foundation-based models trained with massive data and compute. As the number of reference images increases, the FVD of our results shows a noticeable decrease. This improvement can be attributed to two factors: multiple reference frames offer more portrait information, serving as anchor points that simplify generative distribution modeling thus enhances generation quality; additionally, they substantially improve decoding fidelity of VAE, as will be demonstrated in our ablation study.

Crucially, our method enables real-time online generation on a single GPU, a capability that is not supported by these baseline approaches. 
This underscores the potential of our method as a capable solution for human-centric interactive video generation, offering a distinct path from existing approaches based on large foundation models.

\paravspace
\paragraph{Qualitative evaluations.} We refer readers to the \textit{supplementary videos} for visual comparisons between our method and the previous approaches.

\subsection{Ablation Study}

\paragraph{Reference guidance for video VAE.} 
To evaluate the effectiveness of our reference-guided reconstruction, we train variants that exclude the reference injection. For fairness, only the cross-attention layers in $D_{\text{ref}}$ are removed, while all other components remain identical. We further study the effect of the number of reference images, using the same sampling strategy as in the generation comparison.
Experiments are conducted on VoxCeleb2 and HDTF.

Table~\ref{table:ablation_residual_ref} shows that incorporating reference guidance significantly improves video quality. On VoxCeleb2, the PSNR increases from 29.071 to 31.676 (+2.605 dB), and on HDTF from 28.306 to 32.068 (+3.762 dB) with just one reference image used. Moreover, increasing the number of reference images leads to further gains, clearly validating the effectiveness of our reference-guided video VAE design. 

\paravspace
\paragraph{Causal residual video auto-encoding (CR-VA)} We further evaluate the impact of CR-VA within our framework. Without reference guidance, CR-VA brings a slight improvement in reconstruction quality, as shown in Table~\ref{table:ablation_residual_ref}. As the number of reference images increases, CR-VA consistently outperforms the interpolation-based baselines and further \emph{amplifies the benefits of reference guidance}. For instance, on VoxCeleb2, the PSNR gain from $M=0$ to $M=3$ is 4.375 dB with CR-VA, higher than 3.695 dB without it. On HDTF, the corresponding improvement is 6.696 dB versus 4.843 dB. 
These results indicate that CR-VA and the reference guidance strategy can work synergistically to enhance reconstruction quality.

\section{Conclusion}
In this paper, we present a streamable framework for audio-driven talking portrait video generation, capable of producing arbitrary-length videos in real time. At the core of our approach is a causal video VAE with deep latent compression guided by one or more reference images, which focuses on dynamic motion rather than static appearance. A causal residual video auto-encoding scheme is also introduced to further enhance fidelity. Built on this compact latent space, a blockwise autoregressive Rectified Flow Transformer enables efficient long-range generation and seamless streaming. Experiments demonstrate that our method surpasses or matches the visual quality of prior state-of-the-art diffusion foundation based portrait models while achieving over $25\times$ faster generation.

{
    \small
    \bibliographystyle{ieeenat_fullname}
    \bibliography{main}
}

\clearpage

\lstdefinestyle{pytorchstyle}{
    language=Python,
    basicstyle=\ttfamily\footnotesize,
    keywordstyle=\color{blue}\bfseries,
    commentstyle=\color{green!60!black},
    stringstyle=\color{red},
    showstringspaces=false,
    breaklines=true,
    numbers=left,
    numberstyle=\tiny\color{gray},
    captionpos=b,
    aboveskip=10pt,
    belowskip=10pt,
    escapeinside={(*@}{@*)},
    morekeywords={self, nn, Module, CausalConv3d, PixelUnshuffleShortcut, PixelShuffleShortcut},
}

\setcounter{figure}{0}
\setcounter{table}{0}
\setcounter{algocf}{0}
\setcounter{equation}{0}
\setcounter{section}{0}
\renewcommand{\thefigure}{\Roman{figure}}
\renewcommand{\thetable}{\Roman{table}}
\renewcommand{\thealgocf}{\Roman{algocf}}
\renewcommand{\theequation}{\Roman{equation}}
\renewcommand{\thesection}{\Alph{section}}

\begin{figure*}[t!]
   \begin{center}
      \includegraphics[width=0.85\linewidth]{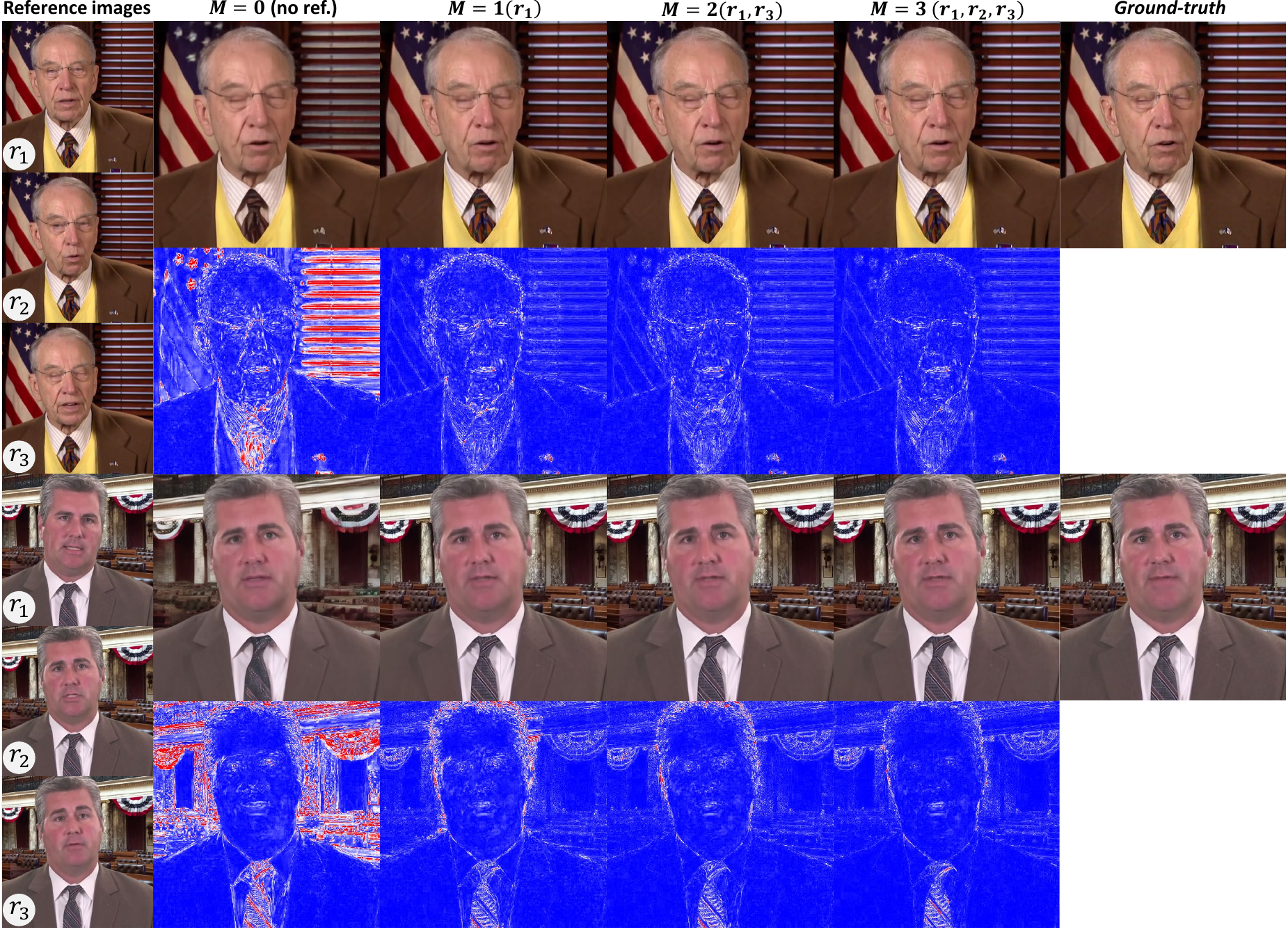}
   \end{center}
   \vspace{-16pt}
      \caption{
      Qualitative results of the reference-guidance module (with CR-VA), with the second and fourth rows showing L2 error maps between predictions and ground truth (red→white→blue indicates high→low error).
      }
   \label{fig:ref_guidance}
\end{figure*}

\begin{figure*}[t!]
   \begin{center}
      \includegraphics[width=0.85\linewidth]{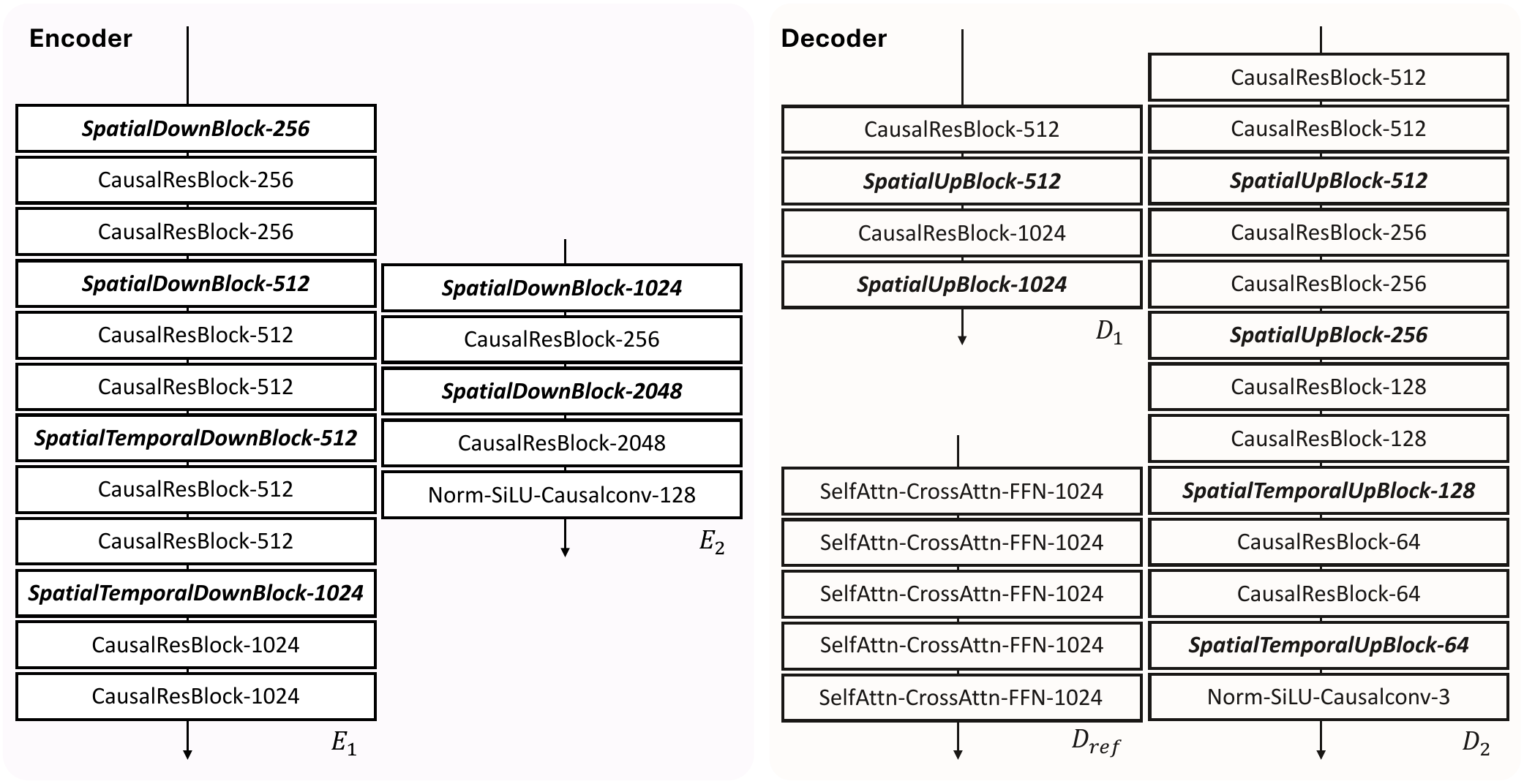}
   \end{center}
   \vspace{-16pt}
      \caption{Network architecture of our causal video VAE. The number after the dash in each block indicates the output channel dimension.}
   \label{fig:network}
\end{figure*}

\section{More Results and Comparisons}
\paragraph{Audio-driven talking head comparison.}
We also compare with several audio-driven talking head methods, including SadTalker~\cite{zhang2023sadtalker} and VASA-1~\cite{xu2024vasa}. These methods rely on motion-disentangled representations to simplify the generation process, but this design choice may also constrain their expressiveness to capture fine-grained details and complex dynamics beyond the head region.
We recropped both the training and test datasets and finetuned our model accordingly. As shown in Table~\ref{table:head_gen}, our method achieves lip-sync ($S_C$, $S_D$) and audio–pose alignment (CAPP) comparable to VASA-1, while producing higher-quality videos, as reflected by FVD.

\begin{table}[h!]
	\centering
	\caption{Quantitative results of audio-driven head video generation on two benchmarks at ${512\times512}$ resolution. Inference speed is measured on a single RTX 4090 GPU.}
    \vspace{-6pt}
	{\scriptsize
    \setlength{\tabcolsep}{1pt}
		\begin{tabular}{c|cccc|cccc|c}
			\toprule
           \multicolumn{1}{c|}{} &  \multicolumn{4}{c|}{HDTF (head)} & \multicolumn{4}{c|}{PortraitOneMin (head)} & Speed \\
            \midrule
			Method  &$S_C\!\!\uparrow$ & $S_D\!\!\downarrow$& \!CAPP$\uparrow$ &$\!\!\text{FVD}_{25}\!\!\downarrow$\!\!
             &$S_C\!\!\uparrow$ & $S_D\!\!\downarrow$& \!CAPP$\uparrow$\! & $\text{FVD}_{25}\!\!\downarrow$ & FPS $\uparrow$\\
			\midrule
            SadTalker & \cellcolor[rgb]{1.00,0.70,0.70}7.497 & \cellcolor[rgb]{1.00,0.70,0.70}7.490 & \cellcolor[rgb]{1.00,0.70,0.70}0.479 & \cellcolor[rgb]{1.00,0.70,0.70}193.843 & \cellcolor[rgb]{1.00,0.70,0.70}6.957 & \cellcolor[rgb]{1.00,0.70,0.70}7.705 &\cellcolor[rgb]{1.00,0.70,0.70}0.413 & \cellcolor[rgb]{1.00,0.70,0.70}270.667 &3.8\\
            VASA-1    & \cellcolor[rgb]{0.70,1.00,0.70}9.489 & \cellcolor[rgb]{0.70,1.00,0.70}5.928 & \cellcolor[rgb]{0.70,1.00,0.70}0.742 & \cellcolor[rgb]{0.87,1.00,0.70}114.863 & \cellcolor[rgb]{0.70,1.00,0.70}8.602 & \cellcolor[rgb]{0.70,1.00,0.70}6.541 &\cellcolor[rgb]{0.70,1.00,0.70}0.679 &\cellcolor[rgb]{0.76,1.00,0.70}119.474 &37.9\\
            \midrule
            $\text{Ours}_{M=1}$ &\cellcolor[rgb]{0.79,1.00,0.70}9.196 &\cellcolor[rgb]{0.74,1.00,0.70}6.024 & \cellcolor[rgb]{0.78,1.00,0.70}0.707 &\cellcolor[rgb]{0.70,1.00,0.70}84.569  &\cellcolor[rgb]{0.85,1.00,0.70}8.182 &\cellcolor[rgb]{0.84,1.00,0.70}6.820  &\cellcolor[rgb]{0.88,1.00,0.70}0.601 &\cellcolor[rgb]{0.70,1.00,0.70}102.468 & 23.3\\
            \bottomrule
		\end{tabular}
	}
	\label{table:head_gen}
    \vspace{-10pt}
\end{table}

\paravspace
\paragraph{Qualitative results for reference-guided video VAE.}
Figure~\ref{fig:ref_guidance} shows reconstruction results for two examples from the HDTF test set under different VAE configurations. The VAE trained without reference guidance ($M=0$) exhibits noticeably higher reconstruction errors. Incorporating a single reference image ($M=1$) substantially improves reconstruction fidelity, reducing errors in static appearance and background regions, and producing sharper, more accurate results. Using multiple reference images further enhances reconstruction quality, consistent with our quantitative results in the main paper.

\paravspace
\paragraph{Qualitative comparisons with prior methods.}
Figure~\ref{fig:talking_portrait_1} and~\ref{fig:talking_portrait_2} present two representative talking-portrait examples along with comparisons to other methods.
Compared with others, our approach demonstrates better identity and appearance consistency with the reference image, more accurate audio–lip synchronization, and more vivid and expressive body and facial dynamics. Moreover, our method supports real-time streaming generation, whereas none of the comparison methods provide this capability.

We provide additional visual results in the \emph{supplementary videos}, including qualitative comparisons with prior methods as well as more examples of our talking-portrait generation. {\textit{We encourage readers to watch these videos for a more comprehensive demonstration of the realism and expressiveness achieved by our method.}

\paravspace
\paragraph{Effects of the classifier-free guidance scale.}
Given the audio and reference images as input conditions, we adopt classifier-free guidance (CFG)~\cite{ho2022classifier} to enhance generation quality. The resulting guided prediction is computed as:
\begin{equation}
\begin{aligned}
G_{\text{cfg}} =& \lambda_a (G_{\text{full}} - G_{a'=\emptyset})
 + \lambda_r (G_{\text{full}} - G_{\mathbf{z}_r=\emptyset}) \\
&\quad - G_{\text{full}} + G_{a'=\emptyset} + G_{\mathbf{z}_r=\emptyset},
\end{aligned}
\label{eq:cfg}
\end{equation}
where $G_{\text{full}}$ denotes the fully conditioned output, and
$G_{a'=\emptyset}$ and $G_{\mathbf{z}_r=\emptyset}$ are obtained by replacing the audio feature $a'$ and the reference latents $\mathbf{z}_r$ with their respective learnable null tokens.
The coefficients $\lambda_a$ and $\lambda_r$ control the guidance strength for each condition, with $\lambda_a = \lambda_r = 1$ corresponding to no CFG\footnote{
An equivalent formulation of Eq.~\eqref{eq:cfg} is
$G_{\text{cfg}} = (1+\lambda_a+\lambda_r)\, G_{\text{full}}
- \lambda_a\, G_{a'=\emptyset}
- \lambda_r\, G_{\mathbf{z}_r=\emptyset}$, with $\lambda_a$ and $\lambda_r$ are defined differently. In this formulation, setting $\lambda_a = \lambda_r = 0$ corresponds to no CFG.
}.

\begin{table}[t!]
    \centering
    \caption{Impact of different classifier-free guidance scales on the HDTF test set. The number of reference images is set to one. }
    \vspace{-6pt}
    {\footnotesize
    \begin{tabular}{l|cccc}
        \toprule
        Settings & $S_C\uparrow$ & $S_D\downarrow$ & $\text{FVD}_{25}\downarrow$ \\
        \midrule
        $\lambda_a=1,\lambda_r=1$ &7.476 &7.311  & 101.112 \\
        $\lambda_a=2,\lambda_r=1$ &9.234 &6.065  & 79.677 \\
        $\lambda_a=2,\lambda_r=2$~\textit{{(default)}} &\textit{8.943} &\textit{6.286}  &\textit{62.300} \\
        $\lambda_a=2,\lambda_r=3$ &8.749 &6.409 & 80.183 \\
        $\lambda_a=3,\lambda_r=2$ &9.262 &6.097 & 71.464 \\
        $\lambda_a=3,\lambda_r=3$ &9.024 &6.279 & 81.358 \\
        \bottomrule
    \end{tabular}
    }
    \label{table:ablation_cfg}
\end{table}

Table~\ref{table:ablation_cfg} presents the effects of different CFG scales. Increasing the audio guidance scale $\lambda_a$ improves lip-sync quality, but large values may reduce diversity, as reflected by $\text{FVD}_{25}$ ($\lambda_a=2$ vs $\lambda_a=3$, with $\lambda_r=2$).  Similarly, a higher reference guidance scale $\lambda_r$ strengthens alignment with the reference image, enhancing stability and overall video quality, but large $\lambda_r$ can negatively impact diversity and lip-sync~($\lambda_r=1,2,3$ while $\lambda_a=2$). Based on these observations, we set $\lambda_a = \lambda_r = 2$ as our default setting.

\section{More Implementation Details}
\paragraph{Causal residual video auto-encoding} Figure \ref{fig:code_comparison} provides the PyTorch-style pseudocode for our implementations of causal residual video auto-encoding.
Consistent with the two-step residual encoding paradigm described in the main text, we decouple resolution changes into temporal and spatial operations.
Two key implementation details are highlighted here:
\begin{itemize}
\item \emph{Causality preservation (split-first strategy)}: To strictly maintain the causal nature of the video frames, we employ a ``split-first'' strategy. The first frame is isolated from temporal downsampling/upsampling operations, ensuring that the latent representation of the first frame depends solely on itself and not on future frames.
\item \emph{Parameter-free residual shortcuts}:
The shortcuts perform dimension matching without learnable parameters.
For downsampling, we apply \texttt{PixelUnshuffle3d} to rearrange space/time into channels, followed by channel grouping and \textbf{mean} reduction. For upsampling, we invert this process by channel duplication (\texttt{repeat\_interleave}) followed by \texttt{PixelShuffle3d} to restore space/time.
\end{itemize}

\paragraph{Network architecture.} Details of our proposed causal video VAE structure are presented in Figure~\ref{fig:network}. All downsampling and upsampling blocks are equipped with the causal residual video auto-encoding (CR-VA) module, as described in Section~3 and Fig.~3 in the main paper. Each CausalResBlock consists of two causal convolution layers with an identity residual connection. The VAE encoder, decoder, and generator have 497M, 406M, and 1B params respectively.

\paragraph{Training details of the causal video VAE.}
We adopt a progressive training strategy for the causal video VAE to reduce training computational cost. Specifically, we first train the auto-encoder at a resolution of $256^2$ using 5-frame clips, with a batch size of 512 on 128 H100 GPUs for 400k iterations. The learning rate is gradually decayed from $1\times10^{-4}$ to $1\times10^{-5}$. We then extend the clip length to 17 frames (batch size 256) and 73 frames (batch size 128), training each stage for 80k iterations with a fixed learning rate of $1\times10^{-5}$. Finally, we finetune the model at a resolution of $512^2$ for last two stages over 40k iterations. The full training process takes approximately one week.

\paravspace
\paragraph{Training details of the denoising model.}
For training the denoising network $G$, we first encode the training videos into video latents $\mathbf{z}$ and reference latents $\mathbf{z}_r$. We then sample segments of 32 latent frames together with 1–3 reference latents for training. To enable CFG, reference latents and audio inputs are randomly dropped with a probability of 0.1. The model is trained with a batch size of 256 on 128 H100 GPUs for 250k iterations, which takes roughly three days.

\paravspace
\paragraph{Inference speed analysis.}
The input to our inference pipeline includes an audio signal and one or more reference images. We assume precomputed image features for generation and decoding. The inference pipeline then consists of three main components: the pretrained audio encoder~\cite{prajwal2020lip}, the rectified flow transformer $G$, and the video decoder $D$. Using the longest KV window, the combined execution of the audio encoder and $G$ (with 12 denoising steps by default) takes $t_G=0.288$ seconds to generate 4 video latent frames (corresponding to 16 video frames). Decoding the 16 frames with $D$ takes $t_D=0.09$ seconds. Overall, the full pipeline achieves real-time performance, with a total latency of $t_G+t_D=0.378$ seconds and a throughput of $16 / (t_G+t_D) = 42.3$ FPS, as reported in Table 1 of the main paper.

\paragraph{Algorithm overview.}
We provide an overview of the training procedure for the generation model $G$ and the full streaming inference process in Algorithm~\ref{alg:generation_training} and Algorithm~\ref{alg:streaming_video_gen}, respectively.

\section{Limitations}
While our method produces realistic and natural talking portraits, it lacks fine-grained control over specific  attributes (\eg, eye gaze and head pose). This could be addressed by introducing additional motion-conditioning signals during video latent denoising. Moreover, the current framework does not support hand motion or larger-scale body movements. We expect these limitations to be addressed by training on datasets that include broader human regions and more diverse, expressive motion patterns.

\section{Ethics Consideration}
This work aims for enabling real-time and interactive virtual avatars, opening up possibilities for various positive downstream applications.
We emphasize that our method is not intended for use in deceptive or malicious scenarios, such as impersonation, misinformation, or any applications that could cause harm, defamation, or other negative impacts on real individuals. We firmly oppose such misuse and are committed to strictly restricting the use of our method to prevent these risks.

Additionally, as a generative model, our method's performance is influenced by biases present in the training data. Therefore, careful attention should be paid to data collection to ensure balanced and unbiased distributions across factors such as race, gender, age, and language.

\begin{figure*}[t]
    \centering
    \begin{minipage}[t]{0.48\textwidth}
\begin{lstlisting}[style=pytorchstyle, title={\small \textbf{Algorithm 1: Spatial-Temporal Downsampling}}]
def PixelUnshuffleShortcut(x, k_t, k_s, c_out):
    """ Residual Branch: Dimension Matching """
    B, C, T, H, W = x.shape
    # 1. Temporal: Space-to-Channel + Channel Averaging
    x0, xt = x[:, :, :1], x[:, :, 1:]
    # Fold time into channel: (B,C,T',H,W)->(B,C*k_t,T'//k_t,H,W)
    xt = PixelUnshuffle3d(xt, factor=(k_t, 1, 1))
    # Group channels and average back to C
    xt = xt.view(B, C, k_t, -1, H, W).mean(dim=2)
    x = torch.cat([x0, xt], dim=2)

    # 2. Spatial: Space-to-Channel + Channel Averaging
    # Fold space into channel: (B,C,T,H,W)->(B,C*ks^2,T,H//ks,W//ks)
    x = PixelUnshuffle3d(x, factor=(1, k_s, k_s))
    # Group channels and average to c_out
    x = x.view(B, c_out, -1, *x.shape[2:]).mean(dim=2)
    return x

class UnshuffleDownsampleBlock3d(nn.Module):
    def __init__(self, c_in, c_out, k_t, k_s):
        self.conv_t = CausalConv3d(c_in, c_in//k_t, k=(1,3,3))
        self.conv_s = CausalConv3d(c_in, c_out//(k_s**2), k=3)
        self.shortcut = PixelUnshuffleShortcut(...)

    def forward(self, x):
        # Residual Encoding Branch
        h = self.shortcut(x, k_t, k_s, c_out)

        # Main Branch: 1. Temporal (Split-First)
        x0, xt = x[:, :, :1], x[:, :, 1:]
        xt = self.conv_t(xt)
        xt = PixelUnshuffle3d(xt, factor=(k_t, 1, 1))
        x = torch.cat([x0, xt], dim=2)

        # Main Branch: 2. Spatial (All frames)
        x = self.conv_s(x)
        x = PixelUnshuffle3d(x, factor=(1, k_s, k_s))

        return x + h
\end{lstlisting}
    \end{minipage}
    \hfill
    \begin{minipage}[t]{0.48\textwidth}
\begin{lstlisting}[style=pytorchstyle, title={\small \textbf{Algorithm 2: Spatial-Temporal Upsampling}}]
def PixelShuffleShortcut(x, k_t, k_s, c_out):
    """ Residual Branch: Dimension Matching """
    # 1. Spatial: Channel Dup + Channel-to-Space
    repeats = c_out * k_s**2 // x.shape[1]
    x = x.repeat_interleave(repeats, dim=1)
    x = PixelShuffle3d(x, factor=(1, k_s, k_s))

    # 2. Temporal: Channel Dup + Channel-to-Time
    x0, xt = x[:, :, :1], x[:, :, 1:]
    xt = xt.repeat_interleave(k_t, dim=1)
    xt = PixelShuffle3d(xt, factor=(k_t, 1, 1))
    x = torch.cat([x0, xt], dim=2)
    return x

class ShuffleUpsampleBlock3d(nn.Module):
    def __init__(self, c_in, c_out, k_t, k_s):
        self.conv_s = CausalConv3d(c_in, c_out*(k_s**2), k=3)
        self.conv_t = CausalConv3d(c_out, c_out*k_t, k=(1,3,3))
        self.shortcut = PixelShuffleShortcut(...)

    def forward(self, x):
        # Residual Encoding Branch
        h = self.shortcut(x, k_t, k_s, c_out)

        # Main Branch: 1. Spatial
        x = self.conv_s(x)
        x = PixelShuffle3d(x, factor=(1, k_s, k_s))

        # Main Branch: 2. Temporal (Split-First)
        x0, xt = x[:, :, :1], x[:, :, 1:]
        xt = self.conv_t(xt)
        xt = PixelShuffle3d(xt, factor=(k_t, 1, 1))
        x = torch.cat([x0, xt], dim=2)

        return x + h
\end{lstlisting}
    \end{minipage}
    \vspace{-1mm}
    \caption{\textbf{Pseudo-code for Spatio-Temporal Resampling.}
    Both the residual shortcut and the main branch follow the same two-step process: temporal then spatial for downsampling, and spatial then temporal for upsampling.
    The shortcut uses \textbf{PixelUnshuffle3d + channel averaging} (down) or \textbf{channel duplication + PixelShuffle3d} (up) for parameter-free dimension matching, while the main branch uses learnable \textbf{CausalConv3d} for the same purpose. The first frame is excluded from all temporal operations (\textbf{Split-First}) to preserve causality.}
    \label{fig:code_comparison}
\end{figure*}

\clearpage

\begin{algorithm}[t!]
\SetKwInOut{Require}{Require}
\SetKwInOut{Ensure}{Output}
\caption{Training Process of Latent Generation Model with Teacher Forcing}
\label{alg:generation_training}
\Require{Video latents $\mathbf{z}$, reference latents $\mathbf{z}_r$, audio feat. $\mathbf{a}$}
\Require{Trainable parameters: generator network $G$}
\For{each training iteration}{
    \tcp{Step 1: Sample latents and apply mask}
    Sample a window of $N=32$ latent frames $\mathbf{z}_\text{win} \subset \mathbf{z}$;

    Sample $M \in \{1,2,3\}$ reference frames $\mathbf{z}_\text{ref} \subset \mathbf{z}_r$;

    \If{first frame is in $\mathbf{z}_\text{win}$}{
        Add learnable mask token $\mathcal{M}$ to first frame;
    }

    \tcp{Step 2: Sample noises}
    Sample $t \sim \text{LogitNormal}(0,1)$ and Gaussian noise $\epsilon^t$ and compute noisy latent:
    $$
        \mathbf{z}^t = t \cdot \epsilon^t + (1 - t) \cdot \mathbf{z}_\text{win}, \quad
        \mathbf{v}^t = \epsilon^t - \mathbf{z}_\text{win};
    $$
    \tcp{Step 3: Add noises to ground-truth to reduce train-test gap}
    Sample $t' \sim \text{Uniform}[0, 0.7]$ and Gaussian noise $\epsilon^{t'}$ to augment ground-truth to provide KV context:
    $$
        \mathbf{z}_\text{aug} = t' \cdot \epsilon^{t'} + (1 - t') \cdot \mathbf{z}_\text{win};
    $$

    \tcp{Step 3: Prepare network input}
    Compute temporally aligned audio embedding $\mathbf{a'}$;

    Concatenate channel-wise: $\mathbf{z}^t$ and $\mathbf{a'}$;

    Flatten along sequence dim: $[\mathbf{z}_\text{ref}, \mathbf{z}_\text{aug}, \mathbf{z}^t \oplus \mathbf{a'}] \to \mathbf{x}_\text{input}$;

    \tcp{Step 4: Forward network with causal blockwise attention (see Fig.~4)}
    $\hat{\mathbf{v}} = G(\mathbf{x}_\text{input}, t)$;

    Compute loss: $\mathcal{L} = \|\mathbf{v}^t - \hat{\mathbf{v}} \|_2^2$;

    Update $G$ via gradient descent;
}
\Ensure{Trained $G$}
\end{algorithm}

\begin{algorithm}[t!]
\caption{Streaming Talking Portrait Video Generation}
\label{alg:streaming_video_gen}
\SetKwInOut{Require}{Require}
\SetKwInOut{Output}{Output}
\SetKwInput{Hyperparams}{Hyperparams}

\Require{Reference frames $I_r$, Generation network $G$, Decoder $D$, and audio stream $A$}
\Output{Streaming video output $y_i$}
\tcp{Step 1: Precomputation and initialization}
Precompute $\mathbf{f}_{r} = E_1(I_r)$; $\mathbf{z}_r = E(I_r)$; $\_,\text{KV}_r = G(\mathbf{z}_r)$;

Initialize KV list with reference KV: $\text{KV}_{\text{list}} = [\text{KV}_r]$ ;

Set current index $i=0$;

\While{audio stream A is live}{
    Wait until next $A_i = 16$ audio frames are ready;

    Compute Audio Embedding $a'_i$ from $A_i$;

    \tcp{Step 2: Denoising Process}
    \For{$j=\text{denoising steps}~T$ down to $0$}{
        \If{$i = 0$}{
            Add learnable mask token to first frame;
        }
        \If{$j = T$}{
        Sample $\mathbf{z}^{t_j}_i$ from Gaussian noise;
        }
        \tcp{Forward G: predict v and get KV}
        $\mathbf{v}^j_i, \text{KV}^j_i=G(\mathbf{z}^{t_j}_i, t_j, a'_i, \text{KV}_{\text{list}})$

        $\mathbf{z}^{t_{j-1}}_i = \text{SolveODE}(\mathbf{z}^{t_j}_i, \mathbf{v}^j_i, t_j)$;

        $\mathbf{z}^{t_j}_i = \mathbf{z}^{t_{j-1}}_i$;
    }
    \tcp{use $\mathbf{z}^0_i$ to provide context for next window.}
    $\mathbf{z}_{\text{last}}=\mathbf{z}^0_i$;

    \tcp{Step 3: Update KV list}
    Add  $\text{KV}^0_i$ to $\text{KV}_\text{list}$;

    \If{$|\text{KV}_\text{list}| \geq$ \text{maximum KV length per window}}{
        \tcp{Reset $\text{KV}_\text{list}$ per window; Re-feed the last block from the previous window to get KV due to positional encoding.}

        $\text{KV}_{\text{list}} = [\text{KV}_r]$; $\_, \text{KV}_{\text{last}} = G(\mathbf{z}_{\text{last}}, \text{KV}_{\text{list}})$;

         Add $\text{KV}_\text{last}$ to $\text{KV}_\text{list}$;
    }

    \tcp{Step 4: Decoding process. We also reuse cached features from previous frames for causal convolutions, omitted here for brevity.}
    Decode video frames $\mathbf{y}_i=D(\mathbf{z}^0_i, \mathbf{f}_r)$;

    $i=i + 1$;

}
\end{algorithm}
\clearpage

\begin{figure*}[t]
   \begin{center}
      \includegraphics[width=1\linewidth]{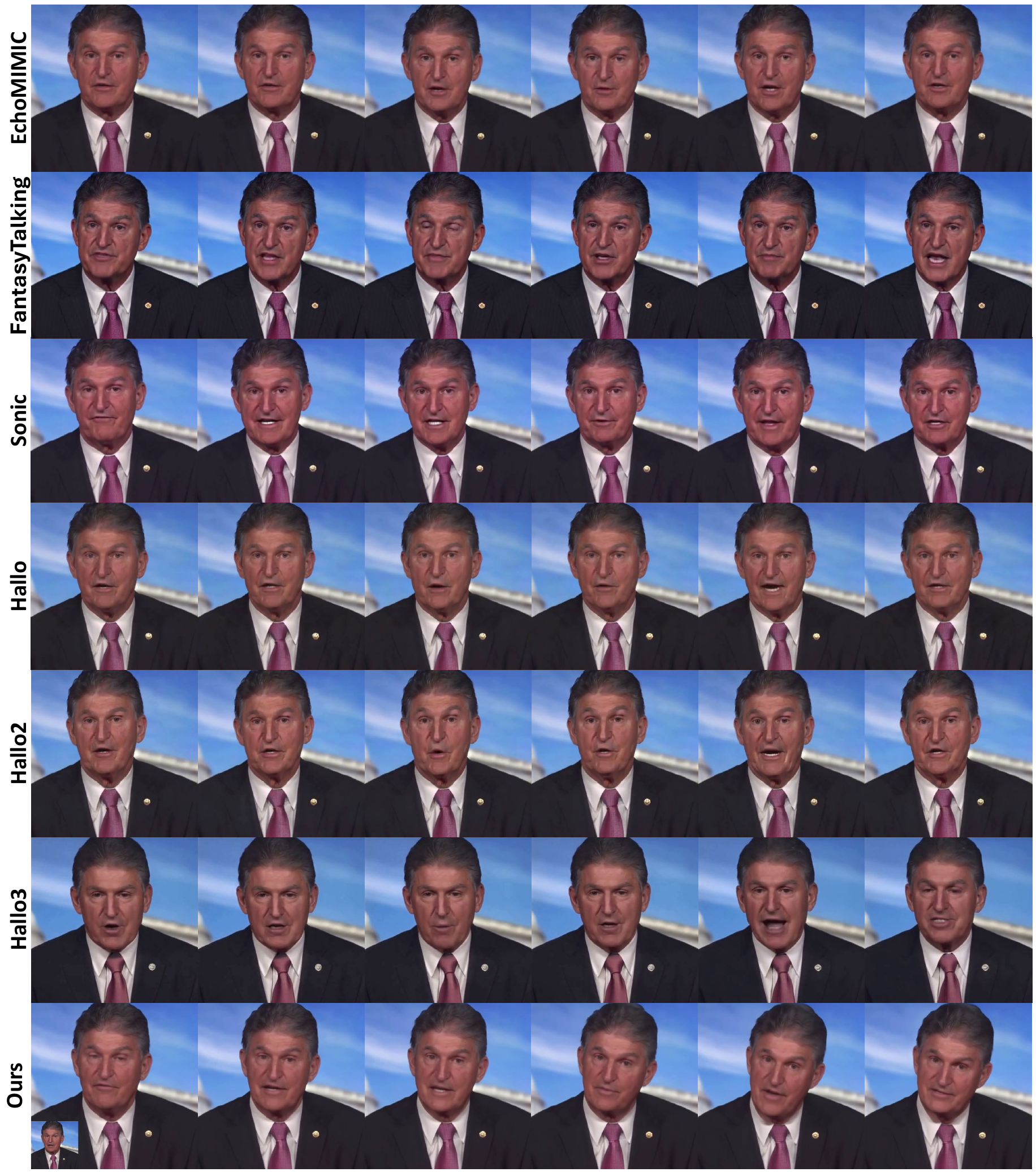}
   \end{center}
   \vspace{-16pt}
      \caption{Talking portrait generation results from different methods using an input audio segment uttering  ``\texttt{face up to that}". The reference image and audio are from the HDTF dataset. Please refer to the \emph{\textbf{supplementary video}} for a more detailed visual comparison.}
   \label{fig:talking_portrait_1}
\end{figure*}

\begin{figure*}[t]
   \begin{center}
      \includegraphics[width=1\linewidth]{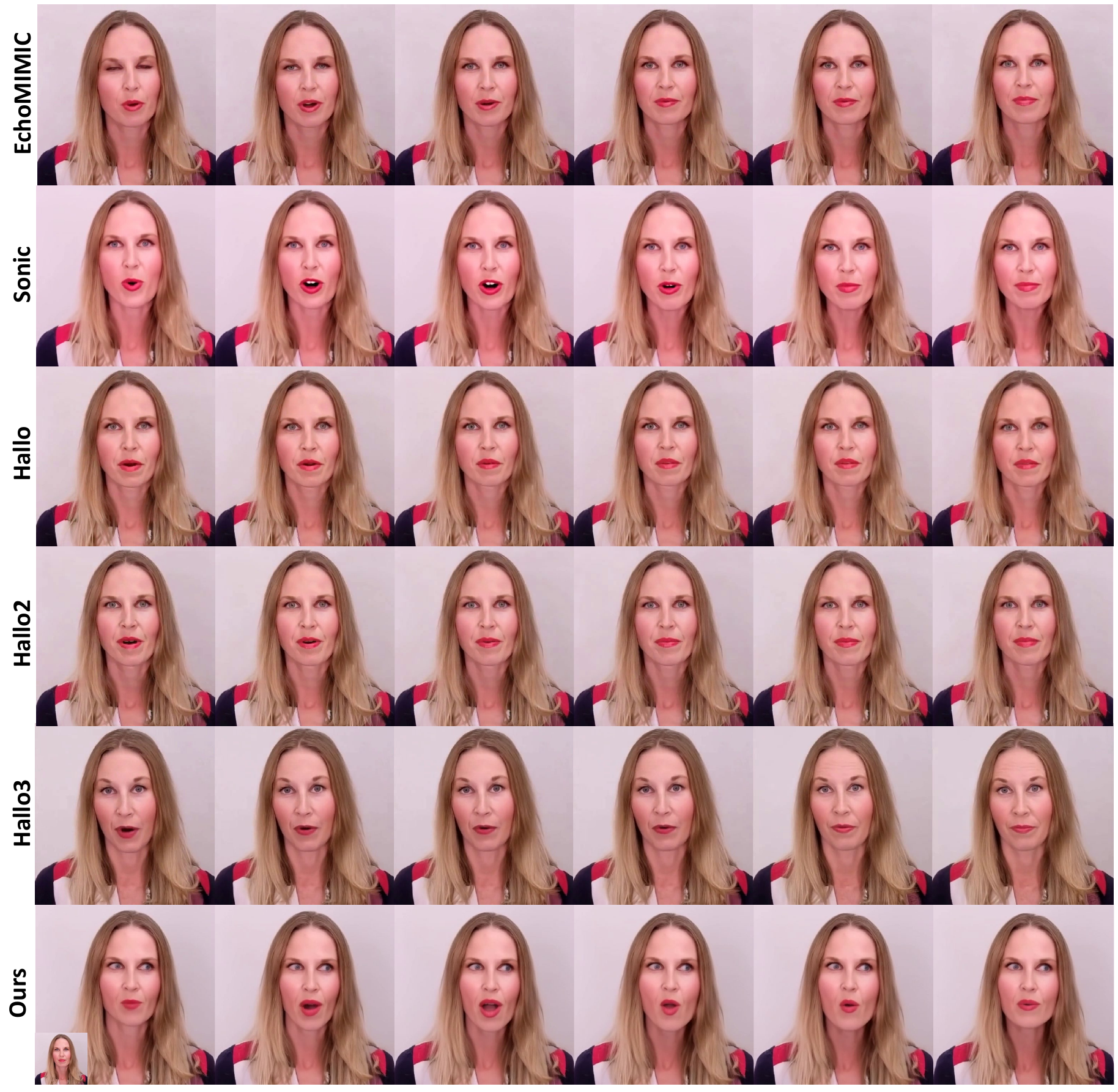}
   \end{center}
   \vspace{-16pt}
      \caption{Talking portrait generation results from different methods using an input audio segment uttering ``\texttt{tomorrow?}". The reference image and audio are from the PortraitOneMin dataset. Please refer to the \emph{\textbf{supplementary video}} for a more detailed visual comparison.}
   \label{fig:talking_portrait_2}
\end{figure*}

\end{document}